\documentclass[11pt]{article}

\usepackage[utf8]{inputenc} % allow utf-8 input
\usepackage[T1]{fontenc}    % use 8-bit T1 fonts
\usepackage[margin=1in]{geometry}
\usepackage{hyperref}       % hyperlinks
\usepackage{url}            % simple URL typesetting
\usepackage{booktabs}       % professional-quality tables
\usepackage{amsfonts}       % blackboard math symbols
\usepackage{nicefrac}       % compact symbols for 1/2, etc.
\usepackage{microtype}      % microtypography
\usepackage{graphicx}
\usepackage[superscript,biblabel]{cite} % superscript citations
\usepackage{array}
\newcolumntype{P}[1]{>{\centering\arraybackslash}p{#1}}
\usepackage{subcaption}
\usepackage{wrapfig}
\usepackage{enumitem}
\usepackage{amsmath}
\usepackage{authblk}
\usepackage{algorithm}
\usepackage{algpseudocode}
\usepackage{parskip} % consistent block paragraphs: no indent, space between paragraphs

\title{TLDChoiceNet: Quantitatively Choosing a Transfer Learning Dataset}

\author[1]{Jing Ning}
\author[1]{James D. Braza}
\affil[1]{Department of Computer Science, Stanford University}
\date{}

\begin{document}

\maketitle

\begin{figure}[t]
\centering
% ---- TEASER / OVERVIEW FIGURE (CVPR-style, top of page 1) ----
% Placeholder: using transfer_learning_architecture.png for now. Replace with a
% dedicated overview illustration (e.g., overview.png) showing the full pipeline:
%   fine-tune dataset + candidate TL datasets/models
%     -> per-class embedding -> learned LoRA-similar reduction
%     -> prediction head -> predicted test-set accuracy.
\includegraphics[width=\linewidth]{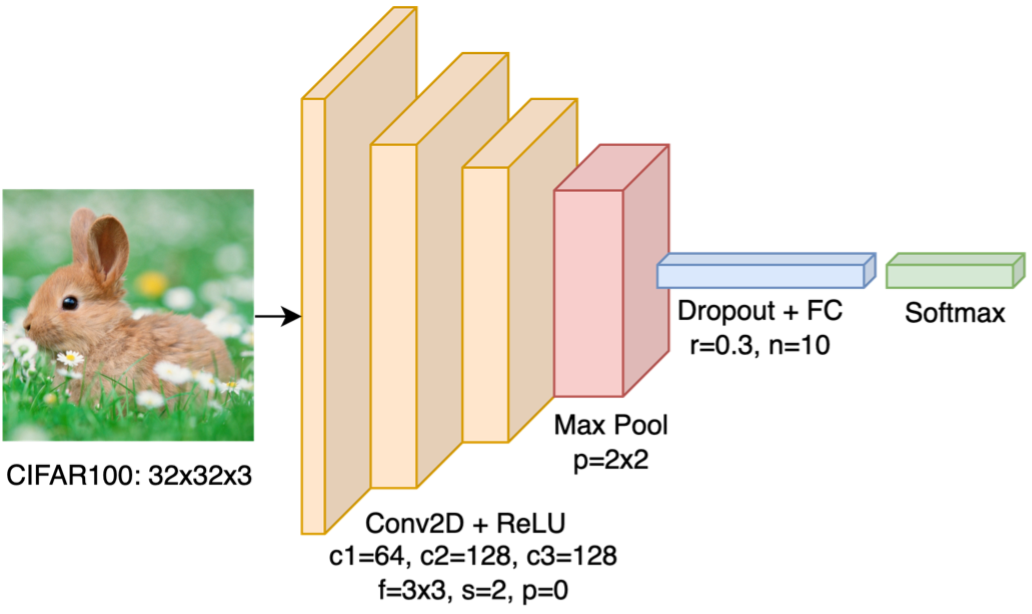}
\caption{\textbf{Overview of TLDChoiceNet.} Given a target fine-tuning dataset
and one or more candidate transfer learning datasets (and/or their pre-trained
models), TLDChoiceNet embeds each input, reduces it through a learned
LoRA-similar layer, and predicts the post-fine-tuning test-set accuracy. This
lets a practitioner choose the best transfer learning dataset \emph{without}
actually fine-tuning on every candidate.}
\label{fig:teaser}
\end{figure}

\begin{abstract}

Transfer learning is particularly useful in settings with limited training data, and within image classification it is common to transfer learn upon massive datasets like ImageNet \cite{imagenet-dataset}, CIFAR-100 \cite{cifar_100_dataset}, or COCO \cite{coco-dataset}.  Qualitatively, it seems a transfer learning dataset should have both more classes and more examples per class than the fine tuning dataset; however, a quantitative method to choose the best transfer learning dataset does not currently exist.  In this paper, we design TLDChoiceNet, a model to choose the best transfer learning dataset given a fine tuning dataset by predicting the test-set accuracy after fine-tuning.  A simple version 1 achieves 0.154 MSE on the test dataset, while a version 2 leveraging an ImageNet pre-trained ResNet50 v2 embedding with per-class information attains a 5X lower MSE of 0.031.  We further design two metrics that enable an unsupervised method of choosing an optimal transfer learning dataset: distribution distance (DD), which linearly regresses against fine-tune accuracy with an R2 of 0.89, and average class correlation (ACC), which improves the R2 to 0.97.  Our results underscore that a dataset's low-level statistics can explain the transfer learning effect, and that using a pre-trained ImageNet can embed different classes further apart in latent feature space.

\end{abstract}

\section{Introduction}

Consider the following scenario.  You have a small dataset for a target task that could be used for fine tuning.  You are aware of three open-source datasets A, B, and C (all with similar example and class count) and pre-trained model weights for A, B, C to use, but only have the resources to fine-tune with one.  Which pre-trained dataset would you choose?  We designed a model TLDChoiceNet which takes datasets A, B, C, and/or pre-trained weights and predicts the test-time accuracy after fine-tuning.  \autoref{fig:teaser} gives a high-level overview of this pipeline.

TLDChoiceNet can also provide:

\begin{itemize}
    \item Pre-trained model to select the best subset class dataset for joint fine-tuning
    \item Summary statistic(s) illustrating why dataset B is desirable.
    \item Insights into whether dataset or pre-trained weights have the most impact on fine-tune performance.
\end{itemize}

We believe answers to these questions could aid the machine learning community at large, and help practitioners speed their transfer learning process in novel areas.

\paragraph{Contributions.} This paper makes the following contributions:
\begin{itemize}
    \item We introduce the \textbf{Transfer Learning Dataset (TLDS)}, a dataset of (transfer learning dataset, transfer-learned model, resultant test accuracy) 3-tuples spanning similar, dissimilar, random, and no-transfer-learning cases.
    \item We design \textbf{TLDChoiceNet (v1 and v2)}, a network with a novel dataset embedding and reduction pipeline that predicts post-fine-tuning test-set accuracy, reducing test MSE 5X from v1 to v2.
    \item We propose two \textbf{unsupervised metrics}, distribution distance (DD) and average class correlation (ACC), that select a transfer learning dataset without any additional training, reaching an $R^2$ of 0.97 against fine-tune accuracy.
    \item We provide \textbf{insights into what is transferred}, showing that pre-trained ImageNet weights embed dissimilar classes further apart in latent feature space.
\end{itemize}

\section{Related Work}

Transfer learning training techniques have been extensively studied, such as selection of fine-tuning layers, fine-tuning robustness, and amenable network architectures \cite{fine-tune-robustness-paper}. Krizhevsky et al. \cite{learning-images-from-layers-paper} showed via training of a Restricted Boltzmann Machine and Deep Belief Net that a network is capable of extracting useful features when trained on other image datasets (e.g.\ hand-written digits, faces).  Another study \cite{pre-training-improve-robustness-paper} focused on using pre-training to improve model robustness and uncertainty.

There is also another research area focusing on how to choose the best data subset for joint fine-tuning. Weifeng et al. \cite{selective-joint-fine-tuning-paper} detail a method to choose the subset of images possessing low-level characteristics similar to the target task, and how to jointly fine-tune shared convolutional layers for both tasks.  Some other interesting approaches \cite{identifying-task-groups-paper} seek to optimally group transfer learning tasks for joint fine-tuning by examining gradients.

Some other areas of research focus on analyzing the fundamental similarity of neural networks.  Kornblith et al. \cite{CKA-similarity-network-paper} developed a quantitative metric called centered kernel alignment to determine the similarity between neural networks.  Peters et al. \cite{to-tune-or-not-paper} researched fine-tuning and feature extraction performance, showing strong correlation between closely aligned tasks.  Neyshabur et al.'s work \cite{being-transfered-in-transfer-learning-paper} has a similar objective as our work: discovering what enables a successful transfer, and which part of the network is responsible. They also presented that some transfer learning benefits come from the low-level data statistics of the data, and showed that pre-trained weights remain within the same loss landscape basin after fine-tuning.

In this paper, we use supervised and unsupervised training techniques, and investigate with multiple custom metrics. Moreover, we also provide insights into what is being transferred, and leverage pre-trained model activations as an embedding method.

\section{Method}

\subsection{The ChoiceNet Algorithm}

We first summarize the full TLDChoiceNet v2 forward pass in \autoref{alg:choicenet}; the remaining subsections detail each component.  Version~1 follows the same pipeline with class-agnostic embeddings and a concatenation (rather than additive) combination step.  Let $D_f$ be the fine-tuning dataset and $D_t$ a candidate transfer-learning dataset with transfer-learned model $M_t$.  We write $g(\cdot)$ for the headless ImageNet-pretrained ResNet50~v2 activation map, with $g(x)\in\mathbb{R}^{2048}$, and $\phi_{M_t}(\cdot)$ for the last-Conv2D activations of $M_t$, with $\phi_{M_t}(x)\in\mathbb{R}^{1152}$.

\begin{algorithm}[t]
\caption{ChoiceNet (v2): predict post-fine-tuning test accuracy}
\label{alg:choicenet}
\begin{algorithmic}[1]
\Require Fine-tuning dataset $D_f$ with classes $\mathcal{C}_f$; candidate transfer-learning dataset $D_t$ with classes $\mathcal{C}_t$ and transfer-learned model $M_t$
\Require Learned parameters $\theta = \{A_f,\, A_t,\, W_1,\, w_2\}$
\State \textbf{Fine-tuning dataset embedding:}
\For{each class $c \in \mathcal{C}_f$}
    \State $e_f^{\,c} \gets \dfrac{1}{|D_f^{\,c}|}\sum_{x \in D_f^{\,c}} g(x)$ \Comment{$g(x)\in\mathbb{R}^{2048}$}
\EndFor
\State $E_f \gets$ rows $e_f^{\,c}$ for the $10$ classes with largest $\lVert e_f^{\,c}\rVert_1$ \Comment{$E_f\in\mathbb{R}^{10\times2048}$}
\State \textbf{Transfer-learning embedding:}
\For{each class $c \in \mathcal{C}_t$}
    \State $e_t^{\,c} \gets \dfrac{1}{|D_t^{\,c}|}\sum_{x \in D_t^{\,c}} \phi_{M_t}(x)$ \Comment{$\phi_{M_t}(x)\in\mathbb{R}^{1152}$}
\EndFor
\State $E_t \gets [\,e_t^{\,1};\ \dots;\ e_t^{\,10}\,]$ \Comment{$E_t\in\mathbb{R}^{10\times1152}$}
\State \textbf{Learned reduction (LoRA-similar):}
\State $R_f \gets E_f A_f,\qquad R_t \gets E_t A_t$ \Comment{$A_f\in\mathbb{R}^{2048\times256},\ A_t\in\mathbb{R}^{1152\times256}$}
\State $H \gets R_f + R_t$ \Comment{$H\in\mathbb{R}^{10\times256}$; \, v1 uses $H=[R_f \,\Vert\, R_t]$}
\State \textbf{Prediction head:}
\State $\hat{y} \gets w_2^{\top}\,\mathrm{Dropout}_{0.3}\!\big(\mathrm{ReLU}(W_1\,\mathrm{vec}(H))\big)$
\State \Return predicted test-set accuracy $\hat{y}$
\end{algorithmic}
\end{algorithm}

The learnable parameters are $\theta = \{A_f, A_t, W_1, w_2\}$, and the model predicts $\hat{y} = \text{ChoiceNet}_{\theta}(D_f, D_t, M_t)$.  Given a TLDS of $N$ tuples $\{(D_f, D_t^{(n)}, M_t^{(n)}, y_n)\}_{n=1}^{N}$, where $y_n$ is the observed post-fine-tuning test accuracy, we train by minimizing the mean-squared error

\begin{equation*}
\theta^{\star} = \arg\min_{\theta}\ \frac{1}{N}\sum_{n=1}^{N}\big(\text{ChoiceNet}_{\theta}(D_f, D_t^{(n)}, M_t^{(n)}) - y_n\big)^2,
\end{equation*}

using the Adam optimizer with a learning rate of $10^{-4}$ for up to 150 epochs, with early stopping (patience of 20 epochs, keeping the best weights).

\subsection{Transfer Learning Dataset Creation}

\begin{figure}[htb]
\centering
\begin{subfigure}{0.44\textwidth}
    \centering
    \includegraphics[width=1.0\linewidth]{transfer_learning_architecture.png}
    \caption{TransferModel Architecture}
    \label{fig:tl-model-architecture}
\end{subfigure}
\begin{subfigure}{0.55\textwidth}
    \centering
    \includegraphics[width=1.0\linewidth]{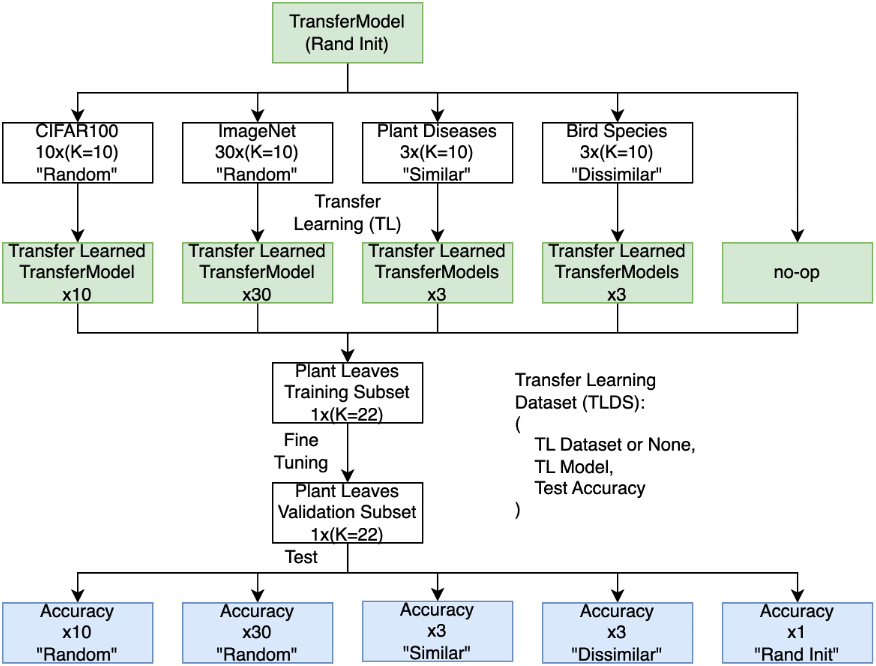}
    \caption{Dataset Generation Process}
    \label{fig:tlds-generation}
\end{subfigure}
\caption{The Transfer Learning Dataset (TLDS) was created with a custom CNN (``TransferModel'') that was repeatedly trained in a full-stack transfer learning, fine-tuning, and testing pipeline.}
\label{fig:tlds-overview}
\end{figure}

We created a novel dataset that we call the Transfer Learning Dataset (TLDS).  This dataset consisted of 3-tuples of (transfer learning dataset, transfer learned model, resultant test accuracy).

First, we created a straightforward CNN called the TransferModel, whose hyperparameters are detailed in \autoref{fig:tl-model-architecture}.  The TransferModel was small to ensure fast training, simple to enable understandable embedding methods, and custom to avoid over-designing for a particular CNN architecture.

We designed the TLDS to have four categories of transfer learning dataset, such that our conclusions were somewhat generalized:
\begin{itemize}
    \item Similar TL dataset: the TL dataset was of plant diseases \cite{plant-village-dataset}, possessing enough classes for three subsets of 10 randomly-sampled classes.
    \begin{itemize}
        \item Note, our fine-tuning dataset was 22-classes of plant leaves \cite{plant-leaves-dataset} from Kaggle.
    \end{itemize}
    \item Dissimilar TL dataset: the TL dataset was of bird species \cite{bird-species-dataset}, possessing enough classes for three subsets of 10 randomly-sampled classes.
    \item Random TL datasets: we randomly sampled classes from CIFAR-100 \cite{cifar_100_dataset} and ImageNet \cite{imagenet-dataset} to create 40 subsets of 10 classes.
    \item No TL dataset: the TransferModel's random initialization was considered the transfer-learned model.
\end{itemize}

We created a script that codified the process detailed in \autoref{fig:tlds-generation}.  Each invocation of the script creates 57 data points, and different seeds enable acquisition of more datapoints.  Here is the gist:
\begin{enumerate}
    \item Store the random initialization of the TransferModel, as an experimental control.
    \item Train (Adam optimizer with learning rate of 0.001 for 15 epochs) upon each TL dataset.
    \item Freeze the TransferModel's first two Conv2D blocks and re-create the head Dense layer. Note the fine-tuning dataset had a different number of classes, standard in transfer learning.
    \item Run fine-tuning (Adam optimizer with learning rate of 0.0001 for 15 epochs) and then test on a validation to attain the test accuracy.
    \item Record the transfer learning dataset subset, post-transfer learning weights, and test accuracy.
\end{enumerate}

\subsection{Transfer Learning Dataset Choice Network: TLDChoiceNet}

\begin{figure}[htb]
\centering
\begin{subfigure}{1.0\textwidth}
    \centering
    \includegraphics[width=1.0\linewidth]{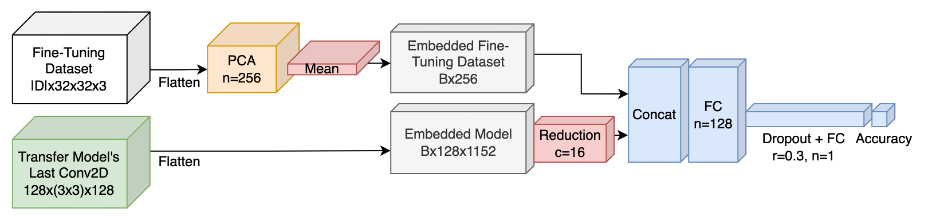}
    \caption{TLDChoiceNet v1 Embeddings and Architecture}
    \label{fig:choicenet-v1-architecture}
\end{subfigure}
\begin{subfigure}{1.0\textwidth}
    \centering
    \includegraphics[width=1.0\linewidth]{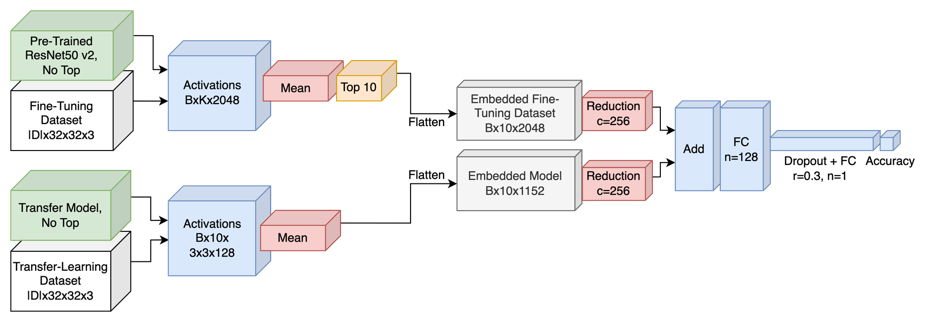}
    \caption{TLDChoiceNet v2 Embeddings and Architecture}
    \label{fig:choicenet-v2-architecture}
\end{subfigure}
\caption{Each version of TLDChoiceNet builds upon the previous.  All figures show the embedding process (left side) and the actual model architecture (right side).}
\label{fig:choicenet-architecture-overview}
\end{figure}

The core design of TLDChoiceNet (see all versions in \autoref{fig:choicenet-architecture-overview}) is a neural network:
\begin{itemize}
    \item Input 1: embedded fine-tuning dataset.
    \item Input 2: embedded transfer learning dataset, transfer learned model, or both.
    \item Output: predicted test-time accuracy, if transfer learning upon the TL dataset.
\end{itemize}

\subsubsection{TLDChoiceNet Training Regime}

Since ChoiceNet performs regression, we use mean-squared error (MSE) as the loss function.  Training was done up to 150 epochs with an Adam optimizer with a learning rate of 0.0001, early-stopping if MSE loss stopped decreasing for 20 epochs (keeping the best weights).

Our TLDS creation process uses 15000 images in the transfer learning dataset and 1000 images in the fine-tuning dataset.  We observed a similar TL dataset tends to lead to higher test-set accuracy.  The less fine-tuning examples present, the more a similar TL dataset stands out.  The more fine-tuning examples, the less effect transfer learning had on test accuracy, which makes sense.  We found 1000 fine-tuning images strikes a reasonable balance between the two extremes.

\subsubsection{Version 1 Architecture}

Our first-pass at ChoiceNet only embedded the transfer-learned model, leaving the transfer learning dataset untouched.  It has a simple design (shown in \autoref{fig:choicenet-v1-architecture}) that provided a baseline performance.

The fine-tuning dataset was embedded in two steps.  First was principal component analysis (PCA) on the entire dataset, to reduce each image from a full RGB image to a vector with 256 elements.  Next, all examples were averaged to lead to a 256-element vector.  The embedding of the transfer-learned model (note: before fine-tuning) was simply each filter in the last Conv2D flattened, which is a matrix of shape $(128, 1152)$.

TLDChoiceNet v1 first performs a LoRA-similar reduction of the embedded TL model.  Instead of two trainable matrices ($A$ and $B$), we just use one trainable $A$ matrix.  This learned reduction further reduces the embedded TL model's dimensionality to $(126, 16)$.  Next, TLDChoiceNet v1's head concatenates both embeddings and passes them through two fully-connected layers ($n = 128, 1$) with a 30\% dropout layer in between.

\subsubsection{Version 2 Architecture}

The 2\textsuperscript{nd} version of ChoiceNet attempted to solve a major issue: both embedding processes ignored class-specifics.  The fine-tuning embedding averaged across all classes, and the transfer learning model's weights general to all classes.  This was fixed with two fundamental changes:
\begin{itemize}
    \item The embedding process was done on a per-class basis.
    \item Post-embedding reduction was done by a trained aspect of ChoiceNet.
\end{itemize}

The transfer learning embedding process begins with the transfer-learned TransferModel's last Conv2D's activations.  We chose activations because it combines learned weights (not specific to class) with dataset (all examples).  We reduced the dimensionality by averaging activations across each class, resulting in a $(10, 1152)$ matrix.  The $10$ corresponds with the number of classes and the 1152 is the flattened size of the last Conv2D.  One final comment is TLDChoiceNet v2 can't use the randomly initialized model since there's not an accompanying dataset to create activations.

The embedding process of the fine-tuning dataset involved using a ResNet50 v2 pre-trained on ImageNet.  We took the model activations before the head layers and averaged on a per-class basis.  Then, since the number of classes was greater than 10, we kept the 10 classes with the highest average absolute value activation.  This left us with a $(10, 2048)$-sized matrix, where the $2048$ comes from the ResNet50 v2's last convolutional layer size.

Now within TLDChoiceNet v2, we used a much larger learned reduction (same LoRA-similar reduction as ChoiceNet v1) to reduce to two $(10, 256)$-sized matrices.  We add both together (instead of concatenation) and share the same head design as TLDChoiceNet v1.  Overall TLDChoiceNet v2 has nearly four times the number of trainable parameters as version 1.

\subsection{Unsupervised Metrics}

\subsubsection{Distance Metric Using Dataset Distribution}

The goal of this experiment was to find metrics which could correlate the performance of transfer learning with fine-tuning task, in an unsupervised fashion.  We designed an evaluation metric called distribution distance (DD) to characterize the difference in location and other variabilities between transfer learning and fine-tune datasets.

DD is a combination of mean, skew and kurtosis between a transfer learning dataset's normalized pixel values and a fine-tune dataset's normalized pixel values.  For skew and kurtosis we take absolute value, since we wanted to treat left-side vs right-side variability equally.  Let us define $i$ as a transfer learning dataset and $j$ as a fine-tuning dataset, with means $\mu_i, \mu_j$, standard deviations $\sigma_i, \sigma_j$, and number of samples $n, m$.  Within a given image, each pixel is denoted ${x_i}_k$ and ${x_j}_k$.  We first define the (absolute) standardized third and fourth moments:

\begin{equation*}
s_i = \left| \frac{1}{\sigma_i^3} \frac{1}{n} \sum_{k=1}^n ({x_i}_k - \mu_i)^3 \right|,
\qquad
\kappa_i = \left| \frac{1}{\sigma_i^4} \frac{1}{n} \sum_{k=1}^n ({x_i}_k - \mu_i)^4 \right|,
\end{equation*}

and analogously $s_j, \kappa_j$ for dataset $j$.  We then define the distance metric $\text{DD}(i, j)$ as the square root of the summed squared errors of mean, absolute skew, and absolute kurtosis:

\begin{equation*}
\text{DD}(i, j) = \sqrt{
    \left( \mu_i - \mu_j \right)^2
    + \left( s_i - s_j \right)^2
    + \left( \kappa_i - \kappa_j \right)^2
}.
\end{equation*}

We then calculate the DD metric for each transfer learning dataset and linearly regress it with test-set accuracy to analyze the correlation.

\subsubsection{Correlation Metric Using a Pre-Trained Network}

In this experiment, instead of just considering dataset distributions, we utilize ImageNet pre-trained weights to yield low-level statistical information on the dataset.

\begin{figure}[htbp]
\centering
\includegraphics[width=0.7\textwidth]{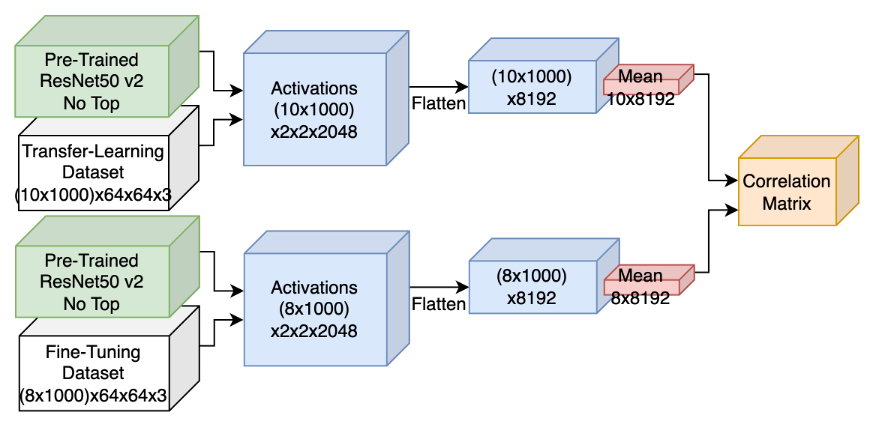}
\caption{Dataset Embedding via ResNet50 v2}
\label{fig:dataset-embedding}
\end{figure}

We do a forward pass through an ImageNet pre-trained headless ResNet50 v2 for all TL dataset images to get the last convolutional layer's output as our embedding latent feature.  Similar to ProtoNet's methodology, we calculate class mean embedding as shown in \autoref{fig:dataset-embedding}.  Each TL dataset is now represented by 10 average class activations.

Now, we can define a new metric: average class correlation (ACC). We have dataset $i$ with $n$ classes and dataset $j$ of $m$ classes.  Let $\text{Cor}_{kl}$ denote the correlation between the per-class embedding of class $k$ in dataset $i$ and class $l$ in dataset $j$.  The average class correlation metric $\text{ACC}(i, j)$ between two datasets is defined as:

\begin{equation*}
\text{ACC}(i, j) = \frac{1}{n\,m} \sum_{k=1}^n \sum_{l=1}^m \text{Cor}_{kl}.
\end{equation*}

We then calculate the ACC metric for each transfer learning dataset and linearly regress it with test-set accuracy to analyze the correlation.

\section{Experiments and Results}

\begin{table}[htb]
\centering
\begin{tabular}{lc}
\toprule
\textbf{Model / Metric} & \textbf{Result} \\
\midrule
TLDChoiceNet v1 --- test MSE & 0.154 \\
TLDChoiceNet v2 --- test MSE & 0.031 \\
Distribution distance (DD) vs.\ accuracy --- $R^2$ & 0.894 \\
Average class correlation (ACC) vs.\ accuracy --- $R^2$ & 0.974 \\
\bottomrule
\end{tabular}
\caption{Summary of key quantitative results.  TLDChoiceNet v2 reduces test MSE by 5X over v1, and the ACC metric explains fine-tune accuracy better than DD.}
\label{tab:summary}
\end{table}

\subsection{TLDChoiceNet Performance} \label{sec:choicenet-performance}

\begin{figure}[htb]
\centering
\begin{subfigure}{0.49\textwidth}
    \centering
    \includegraphics[width=1.0\linewidth]{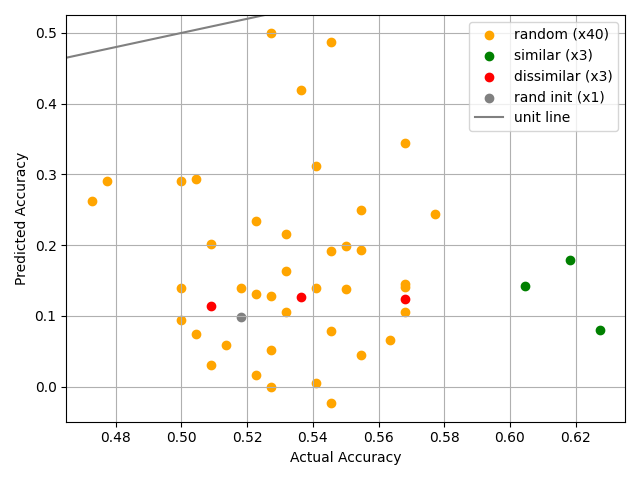}
    \caption{TLDChoiceNet v1 Performance}
    \label{fig:choicenet-v1-performance}
\end{subfigure}
\begin{subfigure}{0.49\textwidth}
    \centering
    \includegraphics[width=1.0\linewidth]{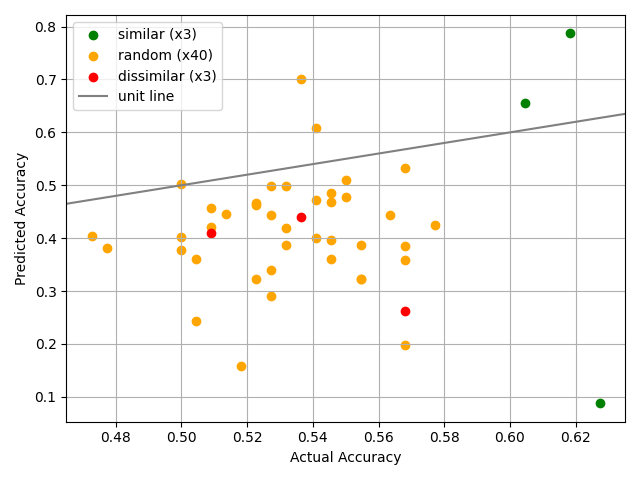}
    \caption{TLDChoiceNet v2 Performance}
    \label{fig:choicenet-v2-performance}
\end{subfigure}
\caption{TLDChoiceNet v1 attained a MSE loss of 0.154 and v2 attained a MSE loss of 0.031 on test datapoints.  Note the test data shared the fine-tuning dataset with the training data.}
\label{fig:choicenet-performance-overview}
\end{figure}

Shown in \autoref{fig:choicenet-performance-overview} we see the results of both TLDChoiceNets.  The ideal network's predicted test-set accuracy matches the actual test-set accuracy, which means all points fall on the unit line.

Before discussing predictions, it's useful to comment on the actual test-set accuracy.  Firstly, the spread from random initialization to similar transfer learned model is only about 15\% accuracy.  Ideally this is a much bigger gap, to underscore the relative effects of initialization, transfer learning, and fine tuning.  Secondly, the dissimilar TL dataset tended to have roughly the same test-set accuracy as the random datasets, which suggests a larger spread of images may not play as big of a role.  To better answer this question, we could redesign the TransferModel or utilize a more challenging test dataset to expose any nuances present.

Not only does TLDChoiceNet v2 have an upgraded architecture, but it has 1{,}147{,}137 trainable parameters compared with v1's 313{,}601 trainable parameters (nearly 4X increase).  At training-time (94 datapoints), both TLDChoiceNet v1 and v2 attained a MSE loss below 0.02.  However, at test time, the difference stands out: v1's MSE loss of 0.154 decreases to 0.031 with v2 (1/5\textsuperscript{th} the size).  In \autoref{fig:choicenet-performance-overview}, the test datapoints comparing actual accuracy with predicted accuracy are plotted.  It's clear that both versions tend to underestimate test-set accuracy, but v2 is vastly closer to the unit line.

MSE numbers listed above were generated from a test subset that shared the same fine-tuning dataset as the training subset.  When testing with both an unseen transfer learning and fine-tuning dataset, TLDChoiceNet v1 does much worse with a MSE of 0.46.  Furthermore, test-set accuracy predictions were off by as much as 80\%.  To handle this removed constraint (shared fine-tuning dataset), we either need a more advanced version of TLDChoiceNet or a more general training set with multiple fine-tuning datasets included.  This outstanding item is also documented in \autoref{sec:future-work}.

Even though there is still room for improvement in a v3 network, the decrease in MSE loss by a factor of 5 at test-time from v1 to v2 is a substantial gain.

\subsection{Low-Level Statistics and Metric Correlation}

By analyzing the histogram of a normalized dataset's pixel distribution, there exists a distribution shift between dissimilar TL dataset classes and fine-tuning dataset classes.  For similar TL dataset classes, the low-level statistics are more similar to fine-tune dataset, as shown in \autoref{fig:data-pixel} (Appendix).

After fitting DD with the test-set accuracy, it can explain the variation in accuracy with a R2 of 0.894 (see \autoref{fig:metric-m} in the Appendix). This shows a strong correlation between the TL dataset's pixel distribution and fine-tune performance.  A lower DD metric thus indicates better test-set accuracy after fine-tuning.  Per-dataset DD values are reported in \autoref{tab:dd-metric} (Appendix).

After embedding TL dataset and fine-tune dataset with a pre-trained ResNet50 v2, we found for 3 groups of similar TL datasets, the ACC metric is 0.501 (see \autoref{fig:corr} in the Appendix).  On the other hand, for dissimilar TL datasets the ACC metric is only 0.276.  This is 45\% lower.  A higher ACC metric could indicate better test-set accuracy after transfer learning.

As illustrated in \autoref{fig:corr-fit} (Appendix), after fitting the correlation metric ACC with the test fine-tune accuracy, it can explain the variation in fine-tune accuracy with a R2 of 0.974.  Thus we can say ACC can better explain the variation on test-set accuracy compared to DD, which only uses dataset distribution.  Per-dataset ACC values are reported in \autoref{tab:acc-metric} (Appendix).

To gain further understanding, we got rid of the pre-trained ResNet50 v2 weights, instead beginning with class means of the dataset's normalized pixels.  We used this to calculate ACC in \autoref{fig:corr}. The average ACC for similar classes is 0.507 and for dissimilar classes it's 0.438, so the difference between similar and dissimilar ACC is 0.06.  However, by adding the ResNet50 v2 pre-trained weights, the average ACC for similar classes is 0.49 and for dissimilar classes it's 0.27. We observe the difference between similar and dissimilar ACC has increased to 0.22.  To summarize this finding: pre-trained weights are able to embed dissimilar class information further apart in latent feature space, providing increased feature separation for use in classification.

\subsection{Pre-Trained Weights Across Fine-Tuning}

To understand how pre-trained weights change across fine-tuning, we measured L2 norm between pre-trained weights and fine-tune weights across all fine-tuning epochs. Shown in \autoref{fig:weight-change} (Appendix), L2 distance changes as a function of the number of epochs at fine-tune time. For a similar TL dataset's pre-trained weights, the L2 distance changes less than a dissimilar TL dataset's pre-trained weights.  This could be due to more shared lower-level features for a similar TL dataset's pre-trained weights compared with a dissimilar TL dataset's weights.

Centered kernel alignment (CKA) is a distance metric invariant to linear transformation and also presents useful information about layers between different models.  We also calculated the CKA as another measure to analyze the feature-space similarity between the pre-trained weights and fine-tuned weights. These results again show that for similar TL dataset, there's more correlation between weights after fine-tuning compared with a dissimilar TL dataset.

\section{Conclusion and Future Work}

In conclusion, we detail TLDChoiceNet v1 and v2, two networks featuring a novel dataset embedding and reduction pipeline, spanning from thousands of examples across many classes to one number: predicted test-set accuracy.  We also designed two metrics, distribution distance (DD) and average class correlation (ACC), that can determine the optimal choice of transfer learning dataset.  DD quantifies the distribution shift between two datasets and ACC proposes correlation between activations can be used as a high-quality substitute for test-set accuracy.

\label{sec:future-work}
Several directions remain for future work:
\begin{enumerate}
    \item One useful activity would be scrutinizing the inner workings of ChoiceNet.  We could answer questions like ``do pre-trained weights, pre-training dataset, or fine-tune dataset play the largest role?'' or ``what aspects of a dataset matter most for test-set accuracy?''  TLDChoiceNet v2 addressed known issues from v1, but it also grew 4X in size.  Perhaps TLDChoiceNet v3 could be vastly smaller but maintain the same performance.
    \item Another area of investigation could be giving ChoiceNet more information on the fine-tuning process.  Perhaps if the fine-tuning is short one pre-trained model is the right choice, but if there's many training epochs or examples, then a different model is the right choice.
    \item We also need to test the generalizability of ChoiceNet to other domains and tasks.  Our problem domain was image classification, so it would be good to test these conclusions in domains without image data or tasks besides classification.  Similarly, we could test with multiple fine-tuning datasets, not just plant leaves.
    \item Analyze and test the ability of other ImageNet pre-trained architectures such as Inception v3 and EfficientNet in further pushing apart feature-space embeddings.
\end{enumerate}

\section*{Acknowledgements}

We thank Chelsea Finn and Daniel Zeng for helpful discussions and feedback, including ideas on ImageNet embedding and measuring weight distance across training.

\bibliographystyle{plain}
\bibliography{reference}

\appendix

\section{Additional Figures and Results}

\subsection{Dataset Distribution: Low-Level Statistics}

\begin{figure}[htb]
\centering
\begin{subfigure}{0.5\textwidth}
    \centering
    \includegraphics[width=5cm]{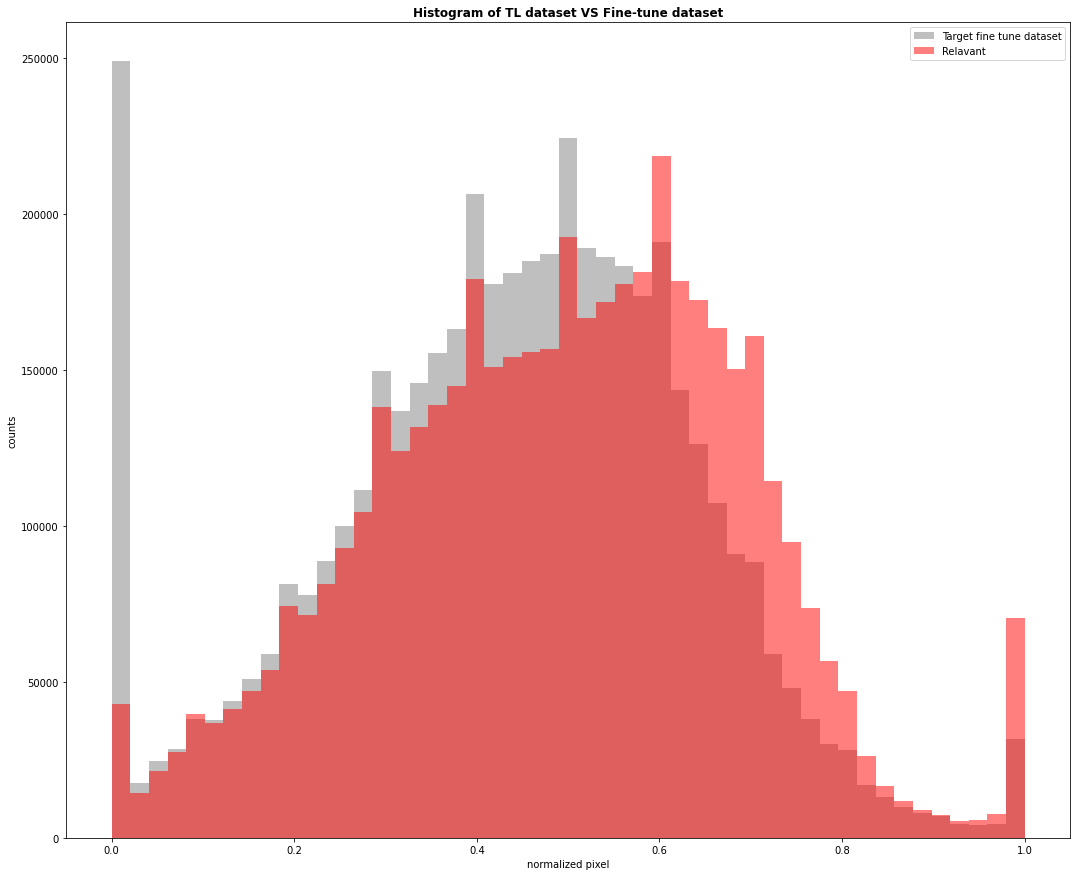}
    \caption{Similar TL Dataset vs Fine-Tune Dataset}
\end{subfigure}
\begin{subfigure}{0.5\textwidth}
    \centering
    \includegraphics[width=5cm]{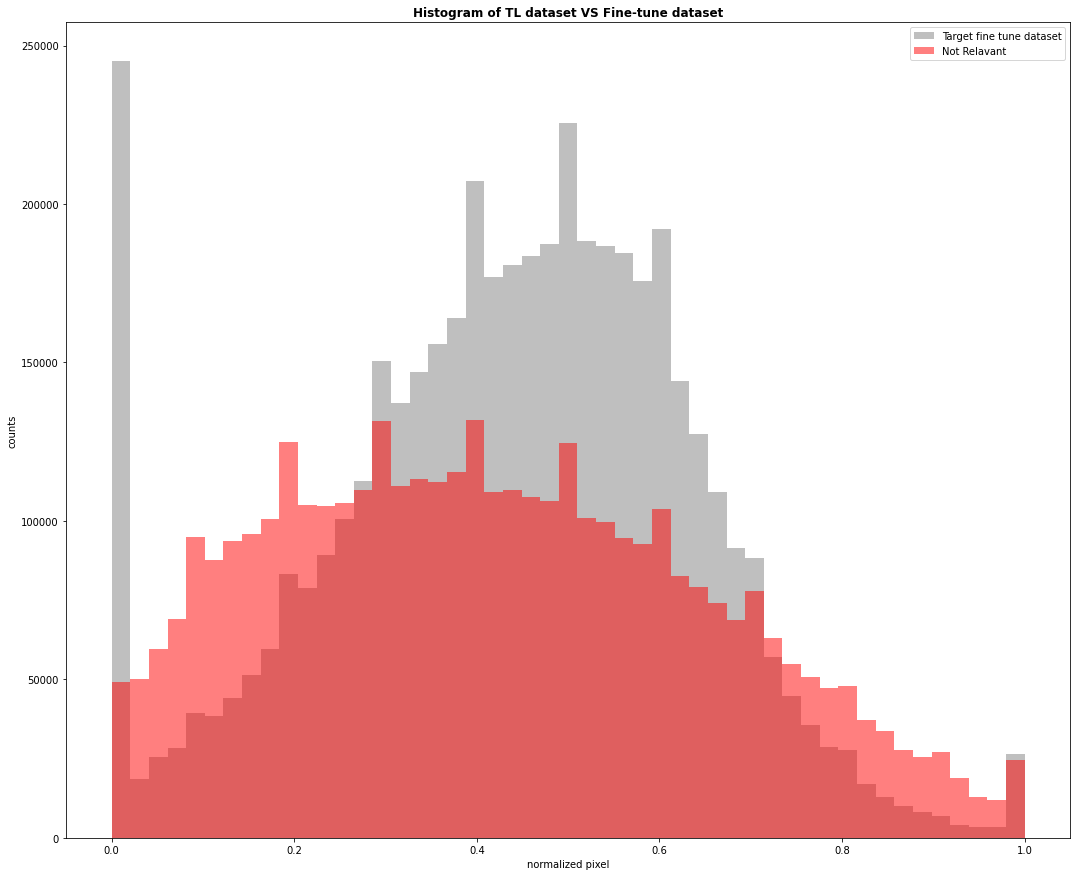}
    \caption{Dissimilar TL Dataset vs Fine-Tune Dataset}
\end{subfigure}
\caption{Histogram of Normalized Pixel Values.}
\label{fig:data-pixel}
\end{figure}

\begin{table}[htb]
\centering
\begin{tabular}{lcc}
\toprule
\textbf{TL Dataset} & \textbf{Distance Metric (DD)} & \textbf{Fine-tune Accuracy} \\
\midrule
Similar -- Model1    & 0.375914939 & 89.0\% \\
Similar -- Model2    & 0.231030915 & 84.5\% \\
Similar -- Model3    & 0.328677488 & 84.5\% \\
Dissimilar -- Model4 & 0.687387568 & 15.5\% \\
Dissimilar -- Model5 & 0.584759946 & 18.5\% \\
Dissimilar -- Model6 & 0.653659486 & 20.3\% \\
Dissimilar -- Model7 & 0.540906724 & 25.4\% \\
Dissimilar -- Model8 & 0.612594293 & 20.3\% \\
Dissimilar -- Model9 & 0.611245781 & 22.4\% \\
\bottomrule
\end{tabular}
\caption{Distance Metric DD vs.\ fine-tune accuracy per TL dataset.}
\label{tab:dd-metric}
\end{table}

\begin{figure}[htb]
\centering
\includegraphics[width=6.0cm]{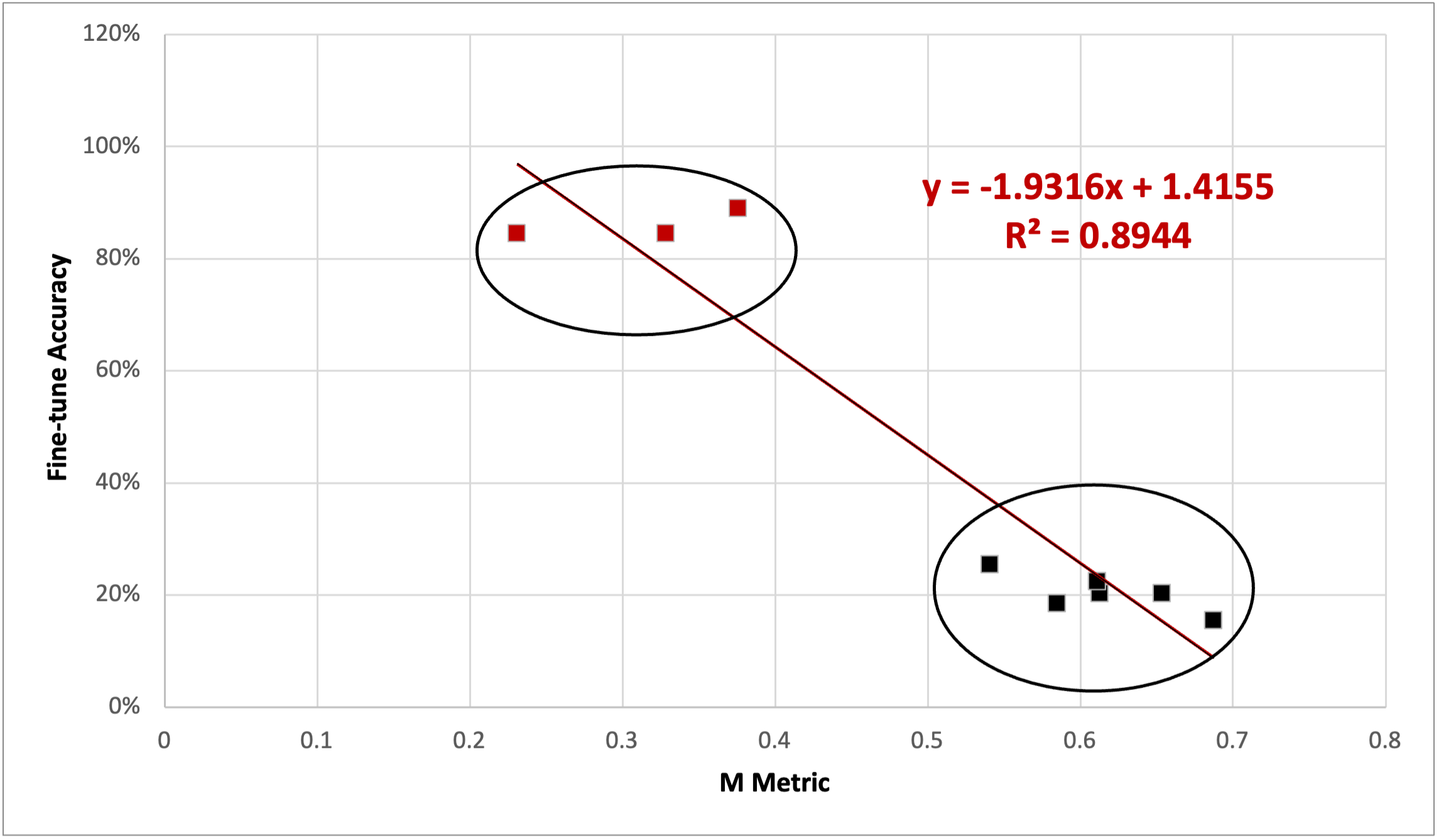}
\caption{Distance Metric DD's Fitting Result.}
\label{fig:metric-m}
\end{figure}

\subsection{Activation Map Correlations}

\begin{figure}[htb]
\centering
\begin{subfigure}{0.5\textwidth}
\includegraphics[width=6.5cm]{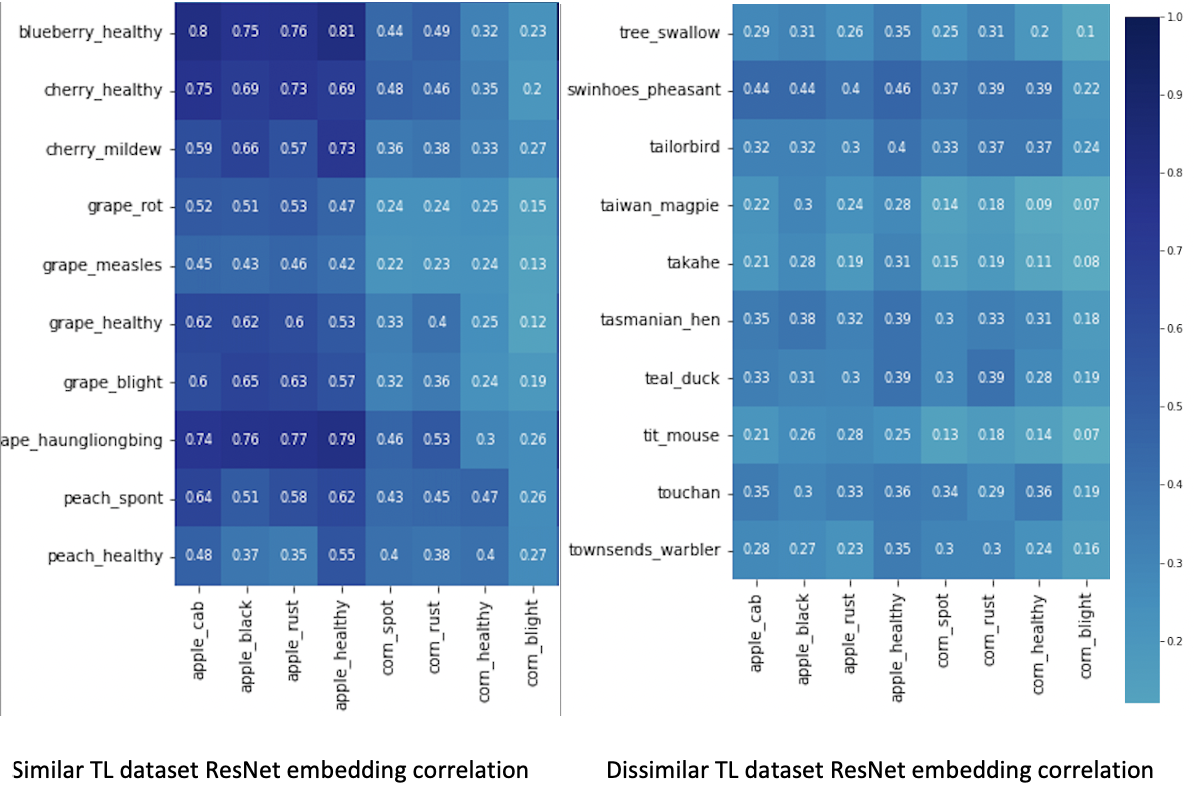}
\caption{ResNet50 v2 Embedding's Correlation Matrix}
\end{subfigure}
\begin{subfigure}{0.5\textwidth}
\includegraphics[width=7cm]{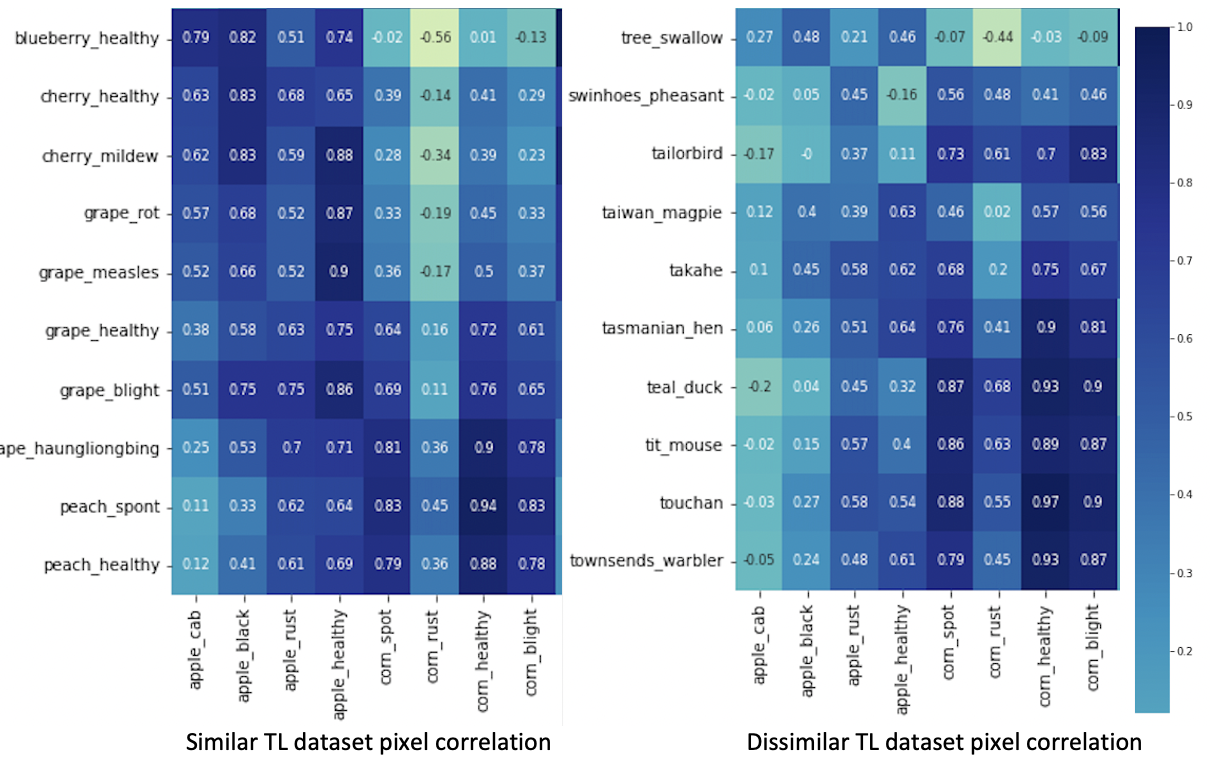}
\caption{Correlation Matrix of Dataset Pixels}
\end{subfigure}
\caption{Correlation Metric ACC vs Test-Set Accuracy.}
\label{fig:corr}
\end{figure}

\begin{table}[htb]
\centering
\begin{tabular}{lcc}
\toprule
\textbf{TL Dataset} & \textbf{Correlation Metric (ACC)} & \textbf{Fine-tune Accuracy} \\
\midrule
Similar -- Model1    & 0.500714286 & 88.99\% \\
Similar -- Model2    & 0.5015      & 84.50\% \\
Similar -- Model3    & 0.49025     & 84.50\% \\
Dissimilar -- Model4 & 0.262       & 15.50\% \\
Dissimilar -- Model5 & 0.2975      & 18.50\% \\
Dissimilar -- Model6 & 0.2865      & 20.30\% \\
Dissimilar -- Model7 & 0.276125    & 25.40\% \\
Dissimilar -- Model8 & 0.252625    & 20.30\% \\
Dissimilar -- Model9 & 0.251625    & 22.43\% \\
\bottomrule
\end{tabular}
\caption{Correlation Metric ACC vs.\ fine-tune accuracy per TL dataset.}
\label{tab:acc-metric}
\end{table}

\begin{figure}[htb]
\centering
\begin{subfigure}{0.5\textwidth}
    \centering
    \includegraphics[width=7cm]{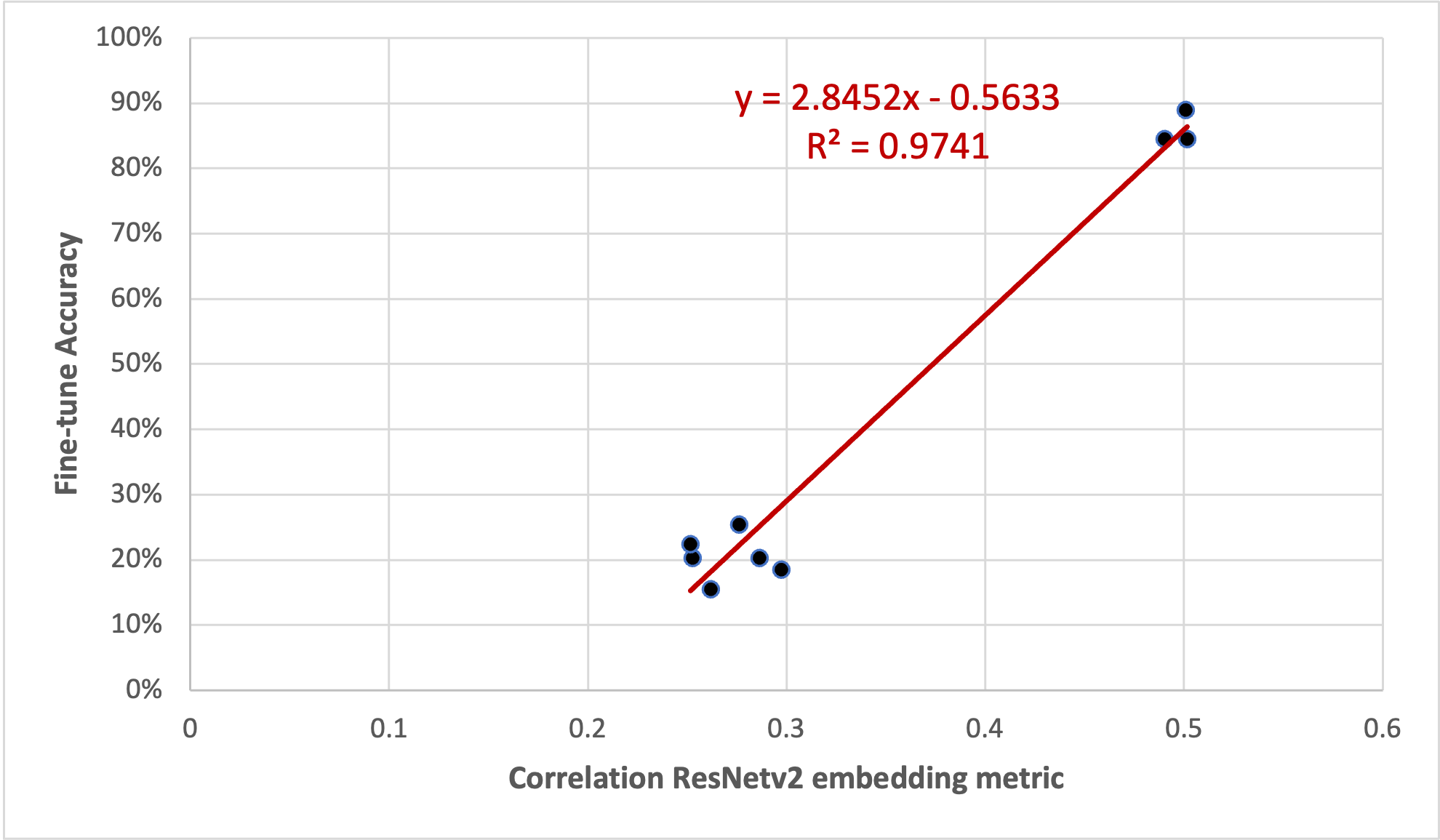}
    \caption{Correlation Metric ACC's Fitting Result}
\end{subfigure}
\caption{Correlation Embedding Metric ACC.}
\label{fig:corr-fit}
\end{figure}

\subsection{Pre-Trained Weights Across Fine-Tuning}

\begin{figure}[htb]
\centering
\begin{subfigure}{0.5\textwidth}
\includegraphics[width=6cm]{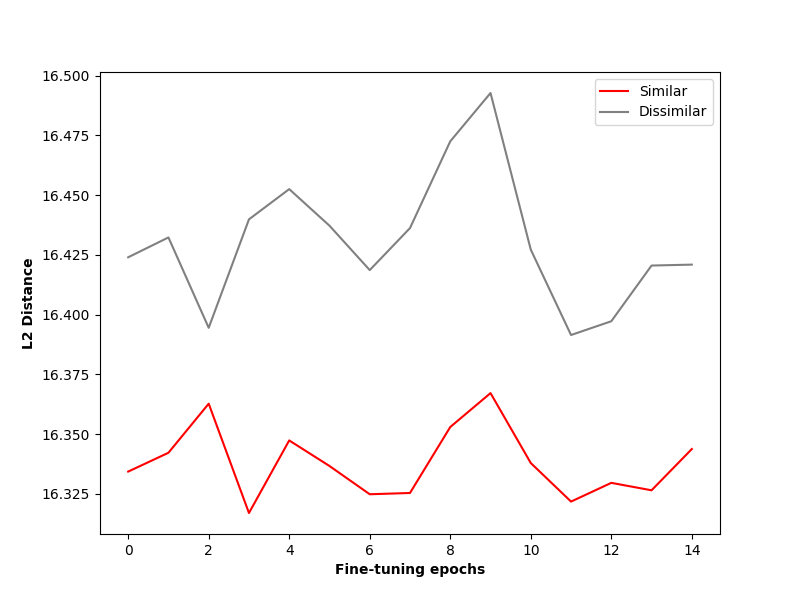}
\caption{Pre-Trained L2 Distance During Fine-Tuning}
\end{subfigure}
\begin{subfigure}{0.5\textwidth}
\includegraphics[width=7cm]{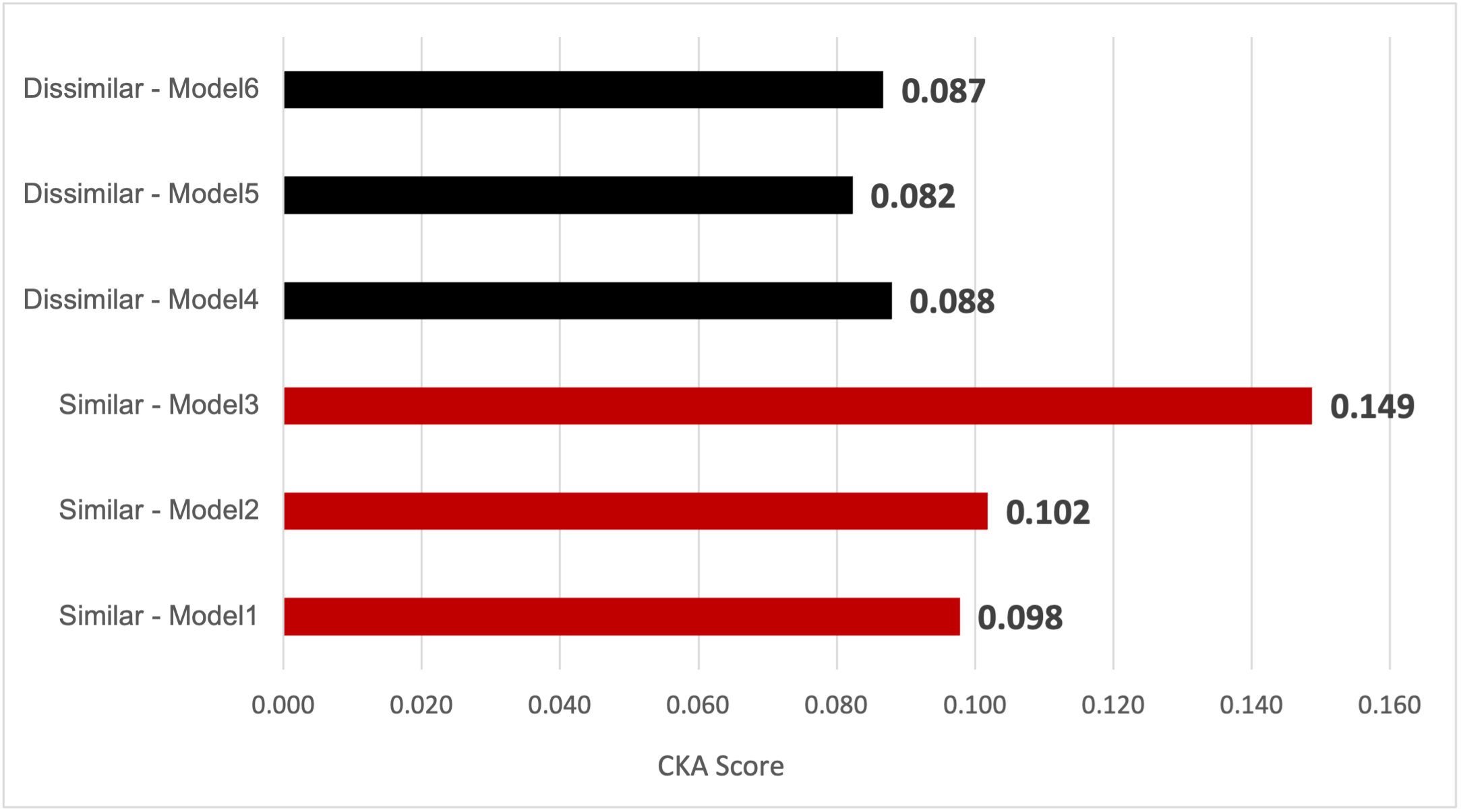}
\caption{Centered Kernel Alignment Metric}
\end{subfigure}
\caption{Weight change and CKA.}
\label{fig:weight-change}
\end{figure}

\section{Source Code}

All source code for the project can be found on GitHub at \href{https://github.com/CDC1688/TLDChoiceNet}{CDC1688/TLDChoiceNet}.

\section{Author Contributions}

J.N.\ designed ChoiceNet v1 and built the transfer-learning and fine-tuning training pipeline used to train it quickly.  J.N.\ also performed the experimental analyses: dataset pixel-distribution analysis, activation maps from a pre-trained ResNet50 v2, the class-correlation metric, monitoring of pre-trained weight-distribution changes during fine-tuning, and the CKA comparison.  J.N.\ created the distribution distance (DD) and average class correlation (ACC) metrics and used them to fit linear regressions against fine-tune accuracy.  J.D.B.\ created functions to load, preprocess, and automatically select subsets of all datasets, and built a comprehensive script to create the TLDS as a TensorFlow Sequence handling the similar, dissimilar, random-initialization, and random-class cases, re-runnable to generate more datapoints per invocation.  J.D.B.\ designed the ChoiceNet v2 embeddings and architecture, building upon the ChoiceNet v1 work, and created scripts to connect ChoiceNet v1 and v2 training with the TLDS.

\end{document}